# マヤ語の台湾先住民(高砂族)語群からの起源

大西耕二 (新潟大・フェロー; ohnishi@bio.sc.niigata-u.ac.jp )

## The origin of Mayan languages from Formosan language group of Austronesian

Koji Ohnishi ( Niigata University (Fellow); ohnishi@bio.sc.niigata-u.ac.jp )

**ABSTRACT:** : Basic body-part names (BBPNs) were defined as body-part names in Swadesh basic 200 words. Non-Mayan cognates of Mayan (MY) BBPNs were extensively searched for, by comparing with non-MY vocabulary, including ca.1300 basic words of 82 AN languages listed by Tryon (1985), *etc*. Thus found cognates (CGs) in non-MY are listed in **Table 1**, as classified by language groups to which most similar cognates (MSCs) of MY BBPNs belong. CGs of MY are classified to 23 mutually unrelated CG-items, of which 17.5 CG-items have their MSCs in Austronesian (AN), giving its *closest similarity score* (CSS), CSS(AN) = 17.5,which consists of 10.33 MSCs in Formosan, 1.83 MSCs in Western Malayo-Polynesian (W.MP), 0.33 in Central MP, 0.0 in SHWNG, and 5.0 in Oceanic [*i.e.*, CSS(FORM)= 10.33, CSS(W.MP) = 1.88, …, CSS(OC)= 5.0 ]. These CSSs for language (sub)groups are also listed in the underline portion of every section of (**§1 ~ §6**) in **Table 1.**  $\chi^2$-test (degree of freedom = 1) using [**Eq 1**] and [**Eqs.2**] revealed that MSCs of MY BBPNs are distributed in Formosan in significantly higher frequency ($P < 0.001$) than in other subgroups of AN, as well as than in non-AN languages. MY is thus concluded to have been derived from Formosan of AN. Eskimo shows some BBPN similarities to FORM and MY.

**[I] 序論と先行研究:** マヤ語族(Mayan family, Mayan = MY)と確実な系統関係を示す言語は知られていない(八杉, 1992; Campbell, 1997)。語族間の遠縁の系統関係の語彙比較による解析は、音韻対応法則を満足する成功例はまだ稀少で(Ohnishi, 2009a,b)、米先住民諸言語(NAmLs)と旧世界言語の系統関係の確実な例としては Dene 語群と Yeniseian の近縁性(Vajda,2010)が広く認められているが、NAmLs の南島語族(Austronesian =AN) 諸分枝との近縁性が最近強く示唆された(Ohnishi, 2012b)。基礎身体部分名称(BBPN)語彙の比較によって、語族間などの遠縁の系統関係が従来の方法よりも遥かに鋭敏な感度で検出できることが報告された(Ohnishi, 2012a,b)。本発表では身体部分名称語彙(BPN) のうち、Swadesh の基礎 200 語彙(安本・本多,1978)の意味を持つ語彙を基礎 BPN(= BBPN = basic BPN)と定義し, MY の BBPN の同祖語(cognate=CG) (意味変化を伴うものを含む)が、どの言語(群)に頻度高く分布するかを解析した。

**[II] 方法:** [1] BBPN としての意味項目(BBP-meaning items = BBPM items)は, Swadesh 基礎 100 語彙 としての 25 項目 [ = {belly (or bowels), blood, bone,breast,ear,egg,fat,feather, foot,hair,hand,head,heart,horn, knee,liver, meat,mouth,neck,nose, saliva(or spit),skin,tail,tongue, tooth }] とそれ以外の 6 項目 [ = {arm, back,breath(or to breathe),leg,lip, wing}] の計 31 項目である。これらの 31 BBPM 項目に対応する意味をもつ語彙を MY より BBPN として選び(1 つの BBPM に複数の BBPN が存在し得る)、その BBPN を MY 以外の諸言語の(基礎的)語彙と比較して(MY BBPN の)CG を検索した。1 つの BBPN に複数の互いに非同祖の CG が存在する場合は、それぞれを異なる CG 項目として解析した。■ [2] 南島語族(Austronesian =AN)との比較には Tryon(1995) "Comparative AN Dictionary" (CAD*)の 80AN 言語の ca.1300 の基礎的語彙や, 台北帝大言語学研究室(小川)(1935)の「付録 単語集」(pp.付 1~付 55) の台湾諸語(高砂族)12 言語の基礎語彙リストを用いた。MY 語彙としては、Kauffman(2003)のマヤ語語源辞典のほ

か、Hoffring (1997), Tozzer (1921), Marin (1995), Vascuez (1980)を適宜参照した。MY 以外の米大陸諸言語語彙としては Swadesh (1954) や Greenberg(1987)の他、いくつかの米大陸言語の各種辞書類や語彙リスト(表 1, §0. (3))を用い、Eurasia 諸言語の語彙としては各言語の辞書類等 (表 1, §0. (3))を用いた。■[3] 最酷似分析(CSA = Closest similarity analysis, Ohnishi, 2012b): 得られた各 CG 項目について、MY 以外の複数言語に互いに同祖の CG が存在するときは、最酷似同祖語(most similar cognate = MSC) がどの言語グループに属するかを調べた、同祖語がただ 1 つしか発見できない場合はそれを MSC とした。各言語群ごとに、言語群 $i$ に MSC が $s_i$ 個の CG 項目について見出だされた場合、最酷似スコア(closest similarity score= CSS) を CSS($i$) = $s_i$ とする。ある MY BBPN について同程度に最酷似する言語群が $m$ 個あるときは、$m$ 個の言語群のそれぞれに $1/m$ の score を加算する形で CSS($i$)を計算し($m= 1~4$)、1 つの MSC についての合計 score が同等に 1 [＝m×(1/m) ]の重みとなるようにした。■[4] 南島語族(AN)については、CSS(AN)の他に、AN の 4 つの subgroup としての、台湾諸語(Formosan =FORM), 西マラヨ・ポリネシア亜族(Western Malayo-Polynesian = W.MP), 中央部 MP(Central MP = C.MP), South Halmahera and West New Guinea (SHWNG), オセアニア語派(Oceanic = OC) の 5 つの亜群ごとに CSS を計算した。■[5] CSA の結果としての語群 Y の亜群ごとの CSS 値が偶然の結果か否かを $\chi^2$-テスト(自由度=1) (Snedecor, 1956; Zar, 1996) で統計評価した。ある語群 Y(=AN)の亜群 X に対する $\chi^2$ 値は, *Freq( i, observed)*= $O_i$ =観察頻度 $_i$, *Freq( i, expected)*= $E_i$ =期待頻度 $_i$ として,

$$\chi_{(X)}^2 = \Sigma_i (Freq( i, observed) – Freq( i, expected))^2 / Freq( i, expected)$$
$$= (O_X – E_X)^2 /E_X + (O_{non-X} – E_{non-X})^2 /E_{non-X} \qquad [Eq.1].$$

であり, $\Sigma_i$ は $i=1$ (=X), $i=2$ (= $Y – X = non-X$ ) に関する和を示す。ここで, (CAD*所収+Ami+Kanakanab の) AN 言語数[$(N_Y=) N_{AN}=82$] は *AN (=Y)よりほぼランダムに選ばれたものと仮定し*、AN の亜群 X の言語数= $n_X$ とすると AN の非 X (non-AN)の言語数 $n_{non-AN} = N_{AN} – n_X$, $O_X = CSS(X), O_{non-X} = CSS(non-X),( O_Y =) O_{AN} = CSS(AN) = O_X + O_{non-X}$ であり,

$$E_x = O_{AN} (n_x /N_{AN} ), E_{non-X} = O_{AN} (n_{non-X}/N_{AN}). \qquad [Eqs 2]$$

である。Eq.1 と Eqs.2 から AN の亜群 X (= FORM, W.MP, etc.)に対する $\chi_{(X)}^2$ 値を得て、Zar(1996)の Table B1(App 13)を用いて，AN 内における MSC の分布を統計評価する。

**[III] 結果と考察:** MY の 31 BBPM 項目に対して、23 CG(同祖語)項目に属する CG が見いだされた。23 CG 項目を MSC の属する言語群ごとに分類し、語源解析結果と CSA 結果と共に表 1 に示す。CSS 値は表中 **§1.~ §6** の各§の冒頭下線部に計算過程を含めて示した。

23 CG 項目に関する CSS の内訳は、23CSS={**CSS(AN)=17.5**, **CSS (Eskimo)= 2.75**, CSS(N.-W.Caucasian)=1, CSS(Turkic)= 0.5, CSS(Mongolic)= CSS(Nahali) = CSS(Tibeto-Burman) =CSS(Macro-Panoan)=CSS(Macro-Ge)= 0.25} = {CSS(AN)= 17.5,CSS(non-AN) = 5.5} であり、殆どの最酷似 CG (MSC)が AN に見られ、Eskimo にもある程度の頻度でみられた。AN の MSCs 17.5 (BBPN)項目の内訳は CSS(AN) =17.5 = **{CSS(FORM)=10.33**, **CSS(MP) =7.16}** = **{CSS(FORM)=10.33**, CSS(W.MP)=1.83, CSS (C.MP) =0.33, CSS(SHWNG)=0, CSS(OC)=5} であり、殆どが FORM に集中してみられた。比較した 81AN 言語の構成は、{5FORM+26W.MP+6C.MP+2SHWNG+41OC}であるから、AN に発見された 17.5CG 項目の 10.33/17.5 = 59.03％ が、FORM のわずか 5 言語 (AN の 5/82 = 6.1%) に集中分布している。

これらの CSS 値などを Eq.1, Eqs.2 に代入して AN の 5 亜群(X=FORM, …, OC)について分析すると、FORM について $\chi_{(FORM)}^2 = 110.7$ を得た。CSS(FORM)が偶然の結果とする帰無仮説は危険率 P < 0.001 で棄却され、従って MY は FORM に最近縁で、FORM に由来することが結論された。**Table 1** の CG リストでは子音対応上の矛盾は特に指摘できず、この結論は妥当である。台湾付近からの直接または間接の米大陸への北廻り移動である。

**[IV] 結論:** マヤ祖語(pMY)は台湾諸語に最近縁で、台湾諸語に起源すると結論できる。もしもアジア大陸の言語を経て pMY に進化したのであれば、その言語は死滅した可能性がかなり高い。

表 1. マヤ語族基礎身体部分名称の同祖語彙の最酷似同祖語彙(MSC)の所属による分類とその語源解析.

**Table 1.** Etymological analysis of cognates of Mayan basic body-part names classified by the belongings of their most similar cognates (MSCs) ]

___

**§0. Abbreviations :** **(1) General:** "A < B", "B > A" = "A has/had been derived from B" ; "A < > B" = "A is (phylo)genetically related to B"; ;"A<<B", "B>>A" = "A has/had been borrowed from B" ; pXX, p-XX = proto-XX ; dial. = dialect ; E.= East(ern), W.= West(ern), N. = North(ern), S.= South(ern), C. = Central

**(2) Language names:** MY =Mayan {Yuc. = Yucatec, (p,s,v,z,m, f,t, *in* Tozzer, 1921) = (Peto, Sotuta, Valladolid, Tizimin, Motul, San Francisco, Tieul, respective ly) } ; Esk = Eskimo (AAY=Alutiiq Alaskan Yupik, CAY= C. Alskan Yupic, CSY= C. Siberian Yupic, GRI = Greenlandic Inuit, NAI = N.Alaskan Inuit, SPI = Seward Peninsula Inuit), AN= Austronesian [FORM= Formosan, MP = Malayo-Polynesian {W.MP = Western MP(PHIL=Philippines, SND=Sundic, SLW=Sulawesi), OC= Oceanic {W.OC = Western OC, ReOC= Remote OC (NCal = New Caledonian, MicN= Micronesian, C.Pacif.= C. Pacific, PolyN = Polynesian )} ] ; Jpn = Japa- nese ; MNG=Mongolic; TbB= Tibeto-Burman; Cauc.= Caucasian

**(3) References:** Bri* = Bricker *et al.,* 1998; CAD* = Tryon, 1995; CED*= Fortescue *et al.*, 1994; DTL*= Öztopçum 1996; EDAL*= Starostin *et al.*, 2003); Hof*= Hofling,1997; IKJ*= Ohno,1990; Kauf* = Kaufman,2003; Marin*= Marin,1995; MTNFT* = 台北帝大言語研, 1935; NCED*= Starostin & Nikolayev, 1994; Toz*= Tozzer, 1921; Vas* = Vásquez, 1980; UEW*= Rédei,1988

**(4)** Language group(s) having most similar cognate(s) (= MSC(s) ) are written in **Bold** type.

___

**§1. Eskimo(エスキモー語)-like:** **{CSS (Esk) = 2 (§1.1.)+ 0.25(§1.2.)+ 0.5(§ 1.3.) = 2.75}**

**§1.1. Most similar to Esk :** 2 items ▪ # **BELLY(腹);** p-Lowland MY *\*naqʼ* "belly" (< *\*na-q*, where *\*na-* < > OC: *na* "stomach"), Chʼortiʔ *nakʼ* "belly", *naʔk* "stomach" (Kauf\*, 338) ||| **Esk: pInuit** *\*na(ž)aq* /GRI *naaq* "belly, abdomen" (CED\*) ||| OC:(ReOC: NCal) Cemuhi *nà* "stomach" ||| Jpn: Old Jpn *na, na-ka* "inside", Mod. Jpn *naka* "inside, belly" (IKJ\*, 964, 966 ; Ohnishi,2009a) ▪ # **SALIVA/ SPIT (唾液,唾):** MY: Itzaj *kʼaʼchiʼ* "saliva, spit" (Hof\*) ||| Esk: **pEsk** *\*qəcɪʀ*"spit" ||| W.MP: (SND) Balinese *mə-kəčuh* "to spit" ▪ **§1.2. Most similar to Esk/FORM/MNG/Nahali :** 1 item {0.25 score for each} ▪ #**EAR (耳):** pMY *\*xikin* ( / ʃikin / ) /Yuc., Itzaj *xikin* / Tacana *xhkyin* "ear" (Kauf\*, 270), (p,v,z,m,f,t) *šikin* "ear" (Toz\*) ( pMY *\*ʃikin* < *\*ʃikɨn* < *\*ʃikun* < *\*ʃiɣun ~ \*siɣun* < > pEsk *\*ciɣun* "ear", FORM: *siku-tu* "shellfish")||| **FORM: Tsou** *sikutu* "shellfish" (< *\*siɣu-tu* < > Esk: CSY *siɣuta* /pEsk *\*ciɣun* //pMY *ʃikin* "ear") ||| **Esk:** CAY *ciun* /SPI *siun* /CSY *siɣuta* "ear" (< **pEsk** *\*ciɣun* "ear" ), GRI *suit* "ear" ; pEsk *\*ciɣtə-quʀ* / CAY *ciutə-quluk* "snail shell", NAI = *siuti-ʀu(q)* "sheshell" (CED\*) (< > GRI *suit* "ear" ) ||| **MNG:** pMNG *\*čiki* /Written Mongolian *čiki(n)* "ear" (EDAL\*,438; YH\*) ||| **Nahali** *cikn* "to hear", *cigam* "ear" (Kuiper,1962) ||| Note: *xikin = šikin* "ear" < *\*šikɨn < \*šiɣɨn* < * *šiɣun* "ear" < *\*šiɣu-n* < > Esk: CAY *ciun* "ear" (< *\*ciɣun* ) ▪ **§1.3. Most similar to Esk/Turkic:** 1 item {1 = Esk 0.5 + Turkic 0.5} ▪ #**BREAST (胸) (of woman):**

pMY *ʔiim /Yuc. iim "breast" | (MY) im (p,s,z), yim (v) "breast" (Toz*) (< *yim ~ *ʔim ) ||| Esk: SPI, NAI immuk "milk" (< *im-muk < *im- "breast (of woman)" + *muk, where *im- < *yim- ~ *ʔim < > MY im, yim ~ pMY *ʔiim "breast (of woman)", and *muk < >AAY muk "teat", muɣ- "to suck at breast".) |||**Turkic: Kazakh** yemshek /**Kyrgys** emchek /**Tatar** imchäk (DTL*) (< *yim-ček < *yim- "breast ?")

**§ 2. Formosan(台湾諸語)-like: {CSS(FORM)= 0.25 (§1.2.) + 9(§ 2.1.)+ 0.5 (§2.2.)+ 0.33(§2.3.) +0.25 (§2.4.)= 10.33 }** ▪ **§ 2.1. Most similar to FORM: 9 items** ▪ # **BLOOD(血):** pMY *kik' / Yuc. k'iʔk' / Itzaj k'ik' / Ch'ol ch'ich' "blood", PQMp kik' "blood", kiik' "hule" (Kauf* 322-324) | qiq (s, m, f ), qiiq (p,v,z) "blood" (Toz*) ( < pMY *k'iik' (= *qiiq )< *qii-q < *qii- < *qɨii- < *qɨi < *quii- ~ *qui < *quʔiʔi ~ *quʔi < *qu-ɟiɟi ~ *qu-ɟi < FORM *quɟiɟiɬʸ ~ *quɟi "red" < *qu-didiɬʸ ~ *qu-di "red" ) ; pMY *kik' (Kauf*) ||| **FORM: Paiwan** ku-ɟiɟiɬʸ (= qu- ɟiɟiɬʸ < *qu-didiɬʸ ) "red" (qu- ɟiɟiɬʸ < qu-ɟiɬʸ < *qu-ɟiɬʸ < *qu-diL ~ *kʷu-diL ) || C.MP: **Dobel** *kʷudi "red" ( < *qu-di ~ *qu-diL "red" ), kʷudu "blood" ▪ #**BONE(骨):** pMY *b'aaq / Itzaj b'ak / MAMo b'aaq "bone", Ixil b'aq "bone" (< *βaq ~ *baq )||| **FORM: Atayal** βaqniʔ "bone" (< *βaq-niʔ ) || W.MP: (PHIL) **Kagayanen** bəkkəg "bone" [ < *bək- kəg < *bək- (<*baq-) "bone" + *-kəg (<*qaɣ ~ *qag ) "rib", where *qaɣ < > (FORM: Atayal) qaɣ "rib". ] ▪ # **FOOT/LEG(足/脚):** pMY *ʔooq / Yuc. ook / Itzaj ok / pCh'olan *ʔok / Chuj yok / Tuzeltai ʔo:k "foot, leg" (*ʔooq < *ʔoʔoq ) "foot, leg" < > FORM: ʔoʔoʔ "leg ~ foot") ; Izaj utan ok / Mopan utaʔn ok / Ch'ortiʔ utajn yok "sole (= chest of foot)" ( ok, yok "foot" ) ||| **FORM: Ami** (dialects: 馬蘭社, 奇密社) ʔoʔoʔ ʔ "foot ~ leg" (< *ʔoʔokʔ ~ *ʔoʔoq ) (From MTFNT*, Appendix p.6) || W.MP: **Gorontaro** ʔoʔoato "foot" (< *ʔoʔo-ato, where *ʔoʔo- < > Ami ʔoʔoʔ ʔ "foot ~ leg" ) ||| Note: Cf. Ami (dial.: 荳蘭社) kokoʔ // Rukai (dial.) koko "leg ~ foot" (From MTFNT*, Appendix p.6) ( koko > *ʔoʔo < > Ami ʔoʔoʔ ʔ // pMY *ʔooq "foot, leg" ) ▪ # **HORN (角):** pMY *ʔuuk'aaʔ "horn" (< *ʔuuk'-aaʔ, where *ʔuuk'- < *ʔuuk' "bovine ~ calf". ) ||| FORM: **Atayal** ūk "calf" ▪ # **HAND(手):** pMaya *q'ab' (= *qab) /Yuc. k'ab' (< *qhabh- ~ *ghabh- < *qabh- ~ *qaβ- ) ||| Quechuan: **Quechua** ccapa "hand" (< *qaba "hand" ) ||| **FORM: Atayal** qaβaʔ "arm" || OC: **Lau** ʔab "hand, arm", **Rotuman** ʔu-hapa "hand" ( < *-hapa < *qaba ) ||| Uralic: p-Finno-Volgaic *käppä /**Finnish** käppä /**Estonian** käpp "hand, paw" (< *käp-pä < *qap- ); **Mordvin** kepe, käpä, "barfuss" (UEW*, 651) ||| IE: pIE *ghabh- "to give, to receive, to seize" ( < MP: *qab-. Ohnishi, 2009b ) ▪ # **MOUTH (口):** pMY *tyiʔ / (MY) chiʔ, tiʔ, tik. tih-ej "mouth" (Kauf*) ; Itzaj chi' "mouse, lip, edge" (Hof*) ; MY (p,v,z,m,f) tši / (s,t) tšii /Yuc. ciʔ "mouth" ||| **FORM**(MTNFT*)**: Ami** tsidaɹ, tsiɭaɹ "mouth" (< *tsi-daɹ) ; **Kanakanab** iβitʂi (< *iβi-tʂi < *iβi- "lip ?" + *-tʂi "mouth", where *-tʂi < *tʂi(i) "mouth" < > Mayan tši(i) "mouth" ) ▪ # **EYE / FACE (目/顔): Tzotzil, Tojol** sat / **Tzeltal** sit "eye, face, fruit" (Kauf*) (< *sat < *sati < > *saptʂi "face" ) ||| **FORM: Tsou** saptʂi "face" (MTFNT*)) ▪ # **EYE/ FACE**: pMY *Haty (H = weak / h / ) ~ *wachi "face ~ eye"/ Itzaj ichi "eye" ( < *watʂi(i) "eye" < *watʂɨ(i)- < *wadɨ(i)- "to see") ||| **FORM: Rukai** (dial.) oatʂɨɨlɨ, wadɨlɨ "to see" (MTFNT*, Appendix p.44) < *wadɨɨlɨ "to see" < *wadɨɨ-lɨ ~ *wadɨ-ɨlɨ (?), where *wadɨ(ɨ)- < > pMY *wachi ~ *Haty "face, eye" ), Rukai wa-ḍə̄lə "to see" (CAD*, #15.510) ( < *wa-ḍɨlɨ "to see") ▪ # **FOOT(足):** MY: **Tzeltol** akan / **Q'anjob'al** aqanej, yaqan / **Akateko** aqaneh / **Mam** qan / **Tacana, Ostuncalco** tqan "foot" (Kauf*, p.346-347) (< pMY *tqan ~ *taqan "foot". Cf. pMY *ʔqan (Kauf*) ) |||

**FORM: Tsou** *tʔaŋo* (CAD*), *tʔaŋŋo* "leg" (*\*tʔaŋ-ŋo < \*tʔan-ŋo*, where *\*tʔan-< \*tqan* ) ▮  §2.2. Most similar to FORM/W.MP: 1 item = {1 = FORM 0.5, W.MP(SLW) 0.5} ▮ **# NECK (首):**   MY: pYukatecan *\*kaal* /Yuc., Mopa *kaal, kal* "neck, throat" (< *\*kāl* "throat ~ to say" ) ||| **FORM: Atayal** *k-um-āl* "to speak, to say" (= *kāl* + *-um-* (infix) ) || **W.MP:(SLW) Konjo** *kalloŋ* "neck" (< *\*kal-loŋ* ? < *\*kal-* "throat" ) ||| **Esk:** pEsk *\*qała-* "to talk", CSY *qałəɣ-* "to speak, to talk, to say" ( > "throat" > Maya *kal* "neck" ? )   ▮

§2.3. Most similar to: FORM/W.MP(PHIL (Yami)) /C.MP: 1 item {1 = FORM 0.33, W.MP(PHIL) 0.33, C.MP 0.33 } ▮ **# HEAD (頭):**   MY: Yuc. (Hocabá dial.) "head, hair", Itzaj *hoʔol* "head", San Luis Jilotepeque *jaluam* (= *haluam*) / Palin *jaloom* (= *haloom*) "head" (Kauf*, pp.274-275), (p,z; m,f,t) *hool'* "head" (Toz*) ;   From these, we can reconstruct pMY *\*hoʔol ~ \*hoʔol'* "head" < *\*hoʔo-l ~ \*hoʔo-l'*, where *\*hoʔo- < \*hoʕo- ~ \*hoᵏo* < > FORM *\*ʔolo ~ \*ʔoᵏo ~ \*qolo* "head" ( < *\*ʔolu ~ \*ʔulo ~ \*qolu < \*ʔulu ~ \*qulu* < pMP *\*qulu* "head" ) ||| **FORM: Rukai** *aolo* (CAD*), *aoro, auro* (MTPNT*) "head" (< *\*a-ulo*, where *\*-ulo* < > Yami *oʁo* / Tagalog *ūlo* // pMY *hoʔol* "head") ; **Paiwan** *qulu* (CAD*), (dialects) *ʔolo, qolo, qoᵏo* (MTPNT*) "head" ( pAN *\*qúluH* (CAD*) "head" > *qulu > qolo ~ \*ʔolo > \*holo > \*hoʁo ~ \*hoʔo* < > pMY *hoʔol* "head" ) ||| **W.MP: (PHIL) Yami** *oʁo* "head" (<*\*olo < \*ʔolo < \*qolo* "head" ), Tagalog *ūlo* / Isnag *ūlu* // (SND) Sundanese *hulu* Sasak *ulu* "head" (< pAN *\*qúluH* ) || **C.MP Buru** *olo-n* "head" ||| Esk: pEsk *\*quliʁ* "upper part", *\*qulə-* "area above"||| Note 1: *ḷ* = / *ḻ* / || Note 2: Cf. pMY *\*hoʔl* "head" (Kauf*). ▮ §2.4. Most similar to: FORM/TbB/ Macro-Panoan/ Macro-Ge: 1 item (0.25 score for each) ▮ **# NOSE (鼻):** MY: Yuc. *niʔ* / Chuj *nhiʔ* "nose" (Kauf*, 314) (< pMY *\*nhiʔ < \*ŋhiʔ ~ \*nihiʔ <\*ŋihiʔ* "nose ~ mucus" < > Atayal *ŋihiʔ* "mucus") [Cf. pMY *\*nhiiʔ* "nose" (Kauf*)] ||| **Macro-Panoan: (Guaicuru) Mocovi** *niih* /(Mataco)Vilela *nihim* "nose" (LAm*, 248) ||| **Macro-Ge:** (Kamakan) **Cotoxo** *nihieko, niiko* /(Mashakali) Kumanasho *nišikoi* "nose" (LAm*, 248) ||| **FORM: Atayal** *ŋihiʔ* "mucus" ||| **TbB: (W.Himalayan) Rangpa** *nhimi* "nose" (< *\*nhi-mi* )

§3. Most similar to W.MP(西マラヨポリネシア亜族): 1 item (= SND 1) {CSS(W.MP)= SLW 0.5(§2.2.) + PHIL 0.33 (§2.3.) + SND 1.0(§3.) = 1.83} ▮ **# BREAST(胸):** MY: (m, f) *tan* "breast" (Toz*) (< AN *taŋ-* ) ||| **W.MP: (SND) Balinese** *taŋkah* "chest" (< *\*taŋ-kah* )

§ 4. Most similar to OC(オセアニア語派): 5 items ={CSS(OC) =W.OC 4 (§4. 1.)+ C.Pacif. 1(§4. 2.) = 5}

§ 4. 1. Most similar to W.OC:   4 items ▮ **# EGG(卵):** p-Greater K'iche'an *\*molo* / Q'eqchiʔ *mol* (Kauf*) ||| OC: **(W.OC) Kaulong** *molmol* "egg" (< *\*mol* ) ▮ **# LIP(唇):** Yuc. *(u) boxel chi'* "lip", *chi'* "mouth", *xel* (= *šel* ) "piece" (Vas*) ( *boxel chi'* < *\*bo-šel + chi'* "mouth", where *\*bo-šel < \*bo-* "mouth" + *šel* "piece" ); Itzaj *chi'* "mouth, lip, edge" (Hof*, 207) ||| **OC: (W.OC) Kaulong** *βo-n* "mouth" (< *\*βo-n < \*bo-* "mouth" ) ▮ **# TOOTH (歯):** pMY *\*kooh* / Yuc. *ko* / Itzaj *koj* (= *koh* ) "tooth" (Kauf*, 260) ( pMY *\*koh ~ \*kooh* < > OC:*\*koh* "to bite" ) ||| OC: **(W.OC) Kaulong** *koh* "to bite" (CAD*, #04.580) ▮ **# TAIL (尾):** pMY *\*nheh* / Yuc. *ne* / pCh'olan *\*neh* /Poptiʔ *nheh* (Kauf*, 312) ( < *\*ŋheh* < > OC: *aŋe* "tail" ?) ||| OC: (W.OC) Mekeo (N.-W. Mekeo dial.) *aŋe* "tail" (< *\*a-ŋe* ?) ▮ § 4. 2. Most similar to ReOC: 1 item ▮ **# FEATHER(羽):** Yuc. *p'ut* "pluma"   (Vas*) (< *\*but ~ \*buti* "feather" < *\*budi* ) ||| OC: **(C.Pacif.) Eastern Fijian** *βuti-* "feather" (< *\*buti* )

**§ 5. Most similar to N.-W. Caucacian (北・西部コーカサス語族):** 1 item {**CSS(N.-W.Cauc.) =1**}

# HEAR (心臓): MY: Itzaj, Mopan *pixan* (=/piʃan /) "soul,spirit", Yuc *pixaʔn* "soul" ||| N.W. Cauc. (W.Cauc.) Abaza psə / Adyghe psa "soul, spirit" (NCED*, 243) (< pW.Cauc.*psa ~ *psə* "soul")

**§ 6. Most similar to other language groups(その他の語群):** CSS(C.MP) = 0.33(§ 2.3.), CSS(Turkic)=0.5 (§1.3.), CSS(MNG)=0.25(§1.2.), CSS (Nahali)=0.25(§1.2.), CSS(TbB) = CSS(Macro-Panoan) = CSS(Macro-Ge) = 0.25 (§2.4.)

______________________________________________________________________